\documentclass[sigconf]{acmart}

\AtBeginDocument{%
  }

\copyrightyear{2026}
\acmYear{2026}
\setcopyright{cc}
\setcctype{by}
\acmConference[LAK 2026]{LAK26: 16th International Learning Analytics and Knowledge Conference}{April 27-May 01, 2026}{Bergen, Norway}
\acmBooktitle{LAK26: 16th International Learning Analytics and Knowledge Conference (LAK 2026), April 27-May 01, 2026, Bergen, Norway}
\acmPrice{}
\acmDOI{10.1145/3785022.3785100}
\acmISBN{979-8-4007-2066-6/2026/04}

\usepackage{multirow}
\usepackage{listings , acmart-taps}
\usepackage{subcaption}
\usepackage{enumitem}
\usepackage{framed}

\newcommand{\Circled}[1]{{\Large \textcircled{\small #1}}}

\begin{document}

\title{Generate-Then-Validate: A Novel Question Generation Approach Using Small Language Models}

\author{Yumou Wei}
\email{yumouw@andrew.cmu.edu}
\orcid{0009-0002-1364-8300}
\affiliation{%
  \institution{Carnegie Mellon University}
  \city{Pittsburgh}
  \country{USA}
}

\author{John Stamper}
\email{jstamper@andrew.cmu.edu}
\orcid{0000-0002-2291-1468}
\affiliation{%
  \institution{Carnegie Mellon University}
  \city{Pittsburgh}
  \country{USA}
}

\author{Paulo F. Carvalho}
\email{pcarvalh@andrew.cmu.edu}
\orcid{0000-0002-0449-3733}
\affiliation{%
  \institution{Carnegie Mellon University}
  \city{Pittsburgh}
  \country{USA}
}

\renewcommand{\shortauthors}{Wei, Stamper and Carvalho}

\begin{abstract}
We explore the use of small language models (SLMs) for automatic question generation as a complement to the prevalent use of their large counterparts in learning analytics research. We present a novel question generation pipeline that leverages both the text generation and the probabilistic reasoning abilities of SLMs to generate high-quality questions. Adopting a ``generate-then-validate'' strategy, our pipeline first performs expansive generation to create an abundance of candidate questions and refine them through selective validation based on novel probabilistic reasoning. We conducted two evaluation studies, one with seven human experts and the other with a large language model (LLM), to assess the quality of the generated questions. Most judges (humans or LLMs) agreed that the generated questions had clear answers and generally aligned well with the intended learning objectives. Our findings suggest that an SLM can effectively generate high-quality questions when guided by a well-designed pipeline that leverages its strengths.
\end{abstract}






\maketitle

\section{Introduction}

The famous phrase ``less is more'' is hardly accurate when describing the current use of large language models (LLMs) in learning analytics and knowledge (LAK) research. A quick, conservative search in the most recent LAK conference (LAK '25) proceedings~\cite{lak2025} shows the field's strong preference for \emph{large} language models: 24 full and short papers contain the keyword ``large language model'', its acronym ``LLM'', or the plural form of both in the \emph{abstract}, out of which 21 papers (87.5\%) also contain the keywords ``GPT'' or ``ChatGPT'' in the \emph{main text}. Including ``LLM'' in the abstract is a strong indicator that LLMs are at least part of the main contribution of these papers, and the high overlap with ``GPT'' suggests that the GPT model family~\cite{gpt3}, characterized by its enormous parameter count (hundreds of billions), is the dominant choice for LLMs in LAK research.

On the one hand, the tendency to use these large language models in LAK research is understandable---after all, these models have demonstrated impressive capabilities in various benchmarks and are readily available through API services. The ability to use these versatile models out of the box allows researchers to focus on higher-level research agendas and rapid research prototyping, which often overshadows the need to deliberate whether such enormous models are indeed of practical necessity or merely a convenient choice.

On the other hand, this prevailing trend to favor large language models also reminds us of an important question: \textbf{should we also explore the potential of \emph{small language models} (SLMs) in LAK research, or at least a synergy between LLMs and SLMs?} Small language models are typically defined as those with substantially fewer parameters (ranging from millions to a few billion). They have been gaining attention for their potential to offer competitive performance while being more resource-efficient~\cite{belcak2025}. In real-world educational settings, where practical applications are subject to resource, technology, and privacy considerations~\cite{kasneci2023,wang2024}, SLMs can provide a more accessible alternative. Having smaller parameter counts not only implies shorter training and re-training times, faster inference speeds, and lower computational costs, but it also makes SLMs more adaptable to specific tasks or domains through training on domain-specific datasets. This adaptability is a significant advantage in educational contexts~\cite{noakes2024}, where domain-specific knowledge is crucial for generating meaningful and coherent responses. In addition, SLMs can be deployed with limited cloud resources or even on local devices, reducing the attack surface for cybersecurity threats and mitigating the privacy concerns associated with sending sensitive educational data to third-party service providers. These advantages make SLMs an attractive option to strengthen the practical impact of LAK research.

One concern about using SLMs may be the quality of their outputs. However, using question generation as a case study, \textbf{we demonstrate that a novel approach using an SLM called Phi-2~\cite{hughes2023} as both an expansive generator and a selective validator can produce questions that human experts rate as high-quality}. Our question generation pipeline begins with Phi-2 generating a large pool of candidate questions based on a given learning objective. After a basic syntactic filtering to remove malformed questions, we ask Phi-2 to validate the remaining questions using its native \emph{probabilistic reasoning} ability and remove questions for which Phi-2 has low confidence in the generated answers. Lastly, we ask Phi-2 to probabilistically choose from all available learning objectives one that best aligns with each question and only retain the questions whose chosen learning objective matches the original.

We conducted two evaluation studies to validate our approach. The first study involved seven human experts, who were asked to answer 64 multiple-choice questions (MCQs) generated by our approach and determine whether each MCQ tests a given learning objective. This design helps us answer the following two research questions (RQs):
\begin{itemize}[leftmargin=*]
    \item RQ-1: How well do human experts and Phi-2 agree on the correct answer to a generated MCQ?
    \item RQ-2: How well do human experts and Phi-2 agree on whether a generated MCQ tests a given learning objective?
\end{itemize}

In addition, we replicated the human evaluation study with Gemini-2.5-Pro~\cite{google2025} as a surrogate judge, allowing us to investigate the following two complementary RQs:
\begin{itemize}[leftmargin=*]
    \item RQ-3: How well does Gemini agree with human experts on the correct answer to a generated MCQ?
    \item RQ-4: How well does Gemini agree with human experts on whether a generated MCQ tests a given learning objective?
\end{itemize}
These two additional RQs help us understand to what extent an advanced LLM can replicate human judgments in evaluating the quality of MCQs generated by an SLM, offering insight into a potential synergy between an SLM for question generation and an LLM for question evaluation. More broadly, our work invites the LAK research community to explore the potential of SLMs as a standalone or complementary tool to LLMs in other educational applications.

\section{Related Work}

Automatic question generation has been an active research area in education. Generating clear and relevant questions at scale is essential for implementing mastery-based learning~\cite{bloom1968} and testing~\cite{collins2019} strategies that are shown to improve learning outcomes and student engagement~\cite{asher2025}. Some early approaches applied rule-based transformations to turn declarative sentences into questions~\cite{heilman2010,adamson2013}. A core idea behind these approaches that inspired our work is ``overgenerate-and-rank''~\cite{walker2001}: generate an abundance of questions and then use a statistical ranking model to select the best ones~\cite{heilman2010}. 

More recent approaches have built on the success of pre-trained language models~\cite{gpt3}. Most language models available for public use today, small or large, are probabilistic language models~\cite{bengio2003}. These models are capable of both generating text (as they are predominantly used today) and evaluating the probability of a given text sequence (a less explored capability in current applications). Most previous work used the \emph{text generation} ability of LLMs to generate questions that are related to domains such as programming~\cite{doughty2024} and math~\cite{bhushan2025}, or that are tuned to different Bloom’s taxonomy levels~\cite{scaria2024}. Some recent work has explored question generation using SLMs fine-tuned on educational datasets~\cite{bulathwela2023,li2025}. In contrast, our work makes synergistic use of both the text generation and \emph{probabilistic reasoning} abilities of an SLM to \emph{generate and validate} questions without any fine-tuning.

One key design choice that obviated the need for fine-tuning in our work was choosing an SLM specifically suitable for education. Recent work has shown that Phi-2~\cite{hughes2023}, a 2.7B-parameter SLM developed by Microsoft, can be used to build an effective knowledge-component extraction algorithm~\cite{wei2025} that outperformed instructional experts. Phi-2 was trained on high-quality textbook-like data~\cite{gunasekar2023}, making it suitable for educational tasks (including question generation). In addition, Phi-2 is openly accessible through the HuggingFace platform~\cite{wolf2020}, allowing for unrestricted access to the next-token probabilities necessary for probabilistic reasoning. Finally, Phi-2 is relatively small (even among SLMs), able to perform fast inference on a single GPU with a modest memory capacity. These three compelling characteristics, along with its demonstrated effectiveness in educational tasks, make Phi-2 an attractive option for building a question generation pipeline that is effective, efficient, and extensible to various local deployment scenarios.

\section{Method}

We built a question generation pipeline that leverages both the text generation and the probabilistic reasoning abilities of an SLM called Phi-2~\cite{hughes2023}. Our pipeline consists of two main stages: \textbf{expansive generation} (Section~\ref{sec: generation}) and \textbf{selective validation} (Section~\ref{sec: validation}), each of which contains multiple steps. The input to the pipeline is a learning objective (LO), a concise statement describing what the generated questions should assess. The output is a set of questions that are likely to align with the given LO, along with their answers and explanations. The overall design of our pipeline reflects an \textbf{abundance} mindset, with which we believe that \textbf{good questions are in the abundance} and that even an SLM can generate high-quality questions if allowed abundant attempts. Therefore, similar to previous work~\cite{heilman2010}, our pipeline named ``Generate-Then-Validate'' begins with a high generation target (e.g., 200 questions per LO) and applies a novel validation approach based on Phi-2's probabilistic reasoning ability to remove low-quality questions.

In what follows, we describe how our pipeline can be instantiated to generate multiple-choice questions (MCQs). However, the pipeline is flexible and can be adapted to generate other types of question, such as true/false or short-answer questions, by adjusting the steps in the generation and validation stages to suit the desired question format. The core idea of ``generate-then-validate'' remains applicable across different question types.

\subsection{Stage 1: Expansive Generation}\label{sec: generation}

\begin{figure*}[t] % Use figure* to span two columns
\begin{lstlisting}[frame=single,basicstyle=\ttfamily\small,showspaces=false,escapechar=\%,caption={A five-step incremental prompt for generating an MCQ. Text in bold represents prompting text that steers generation. A pair of curly braces \{\} enclose variable text dynamically set by the generation program, such as the learning objective used. Any other text is generated by Phi-2, except for the step labels that begin each block.},label={lst: gen_prompt}]
%\Circled{0} \textrm{Seed the generation with a learning objective}% 

%\textbf{The exercises below are designed to test whether a student can}% 
{describe the impacts of human activity on wetlands and mangroves}%\textbf{.}%

%\Circled{1} \textrm{Generate the stem}%

%\textbf{Multiple Choice (best out of }%{4}%\textbf{ options):}%
%\textbf{1.}% What happens when wetlands are drained and converted for agriculture or 
urban development?

%\Circled{2} \textrm{Generate the answer choices}%

%\textbf{a)}% It has no impact on the environment
%\textbf{b)}% It can lead to soil erosion and flooding
%\textbf{c)}% It can increase biodiversity in the area
%\textbf{d)}% It can improve water quality in the area

%\Circled{3} \textrm{Generate the answer}%

%\textbf{Solution:}%
%\textbf{The correct answer is}% b) {It can lead to soil erosion and flooding.}

%\Circled{4} \textrm{Generate the explanation}%

%\textbf{Explanation:}%
When wetlands are drained and converted for  agriculture or urban development, 
the natural water-holding capacity of the area is lost. This can lead to 
soil erosion and flooding, as there is no longer a natural buffer to 
absorb excess water. Wetlands also play an important role in filtering 
pollutants from water, so their loss can lead to decreased water quality.
\end{lstlisting}
\end{figure*}

%\begin{figure*}[t]
%\begin{lstlisting}[frame=single,basicstyle=\ttfamily\small,showspaces=false,escapechar=\%,caption={A five-step incremental prompt for generating an MCQ. Text in bold represents prompting text that steers generation. A pair of curly braces \{\} enclose variable text dynamically set by the generation program, such as the learning objective used. Any other text is generated by Phi-2, except for the step labels that begin each block.},label={lst: gen_prompt},belowcaptionskip=8pt]
%
%
%{describe the impacts of human activity on wetlands and mangroves}
%
%urban development?
%
%When wetlands are drained and converted for  agriculture or urban development, 
%the natural water-holding capacity of the area is lost. This can lead to 
%soil erosion and flooding, as there is no longer a natural buffer to 
%absorb excess water. Wetlands also play an important role in filtering 
%pollutants from water, so their loss can lead to decreased water quality.
%\end{lstlisting}
%\end{figure*}

The goal of this stage is to produce an abundance of candidate questions that can be pruned in the validation stage. We use Phi-2's text generation ability to \emph{incrementally complete a prompt} that guides Phi-2 to build up an MCQ. The full prompt with labeled incremental steps is shown in Listing~\ref{lst: gen_prompt}.

The generation prompt begins with a seeding statement that sets the context and describes the LO used to generate new questions. We recognize that an LO may be communicated in two distinct forms: an action-based statement as a verb phrase (e.g., ``Explain the water cycle'') or a content-based statement describing a fact (e.g., ``The water cycle involves evaporation, condensation, and precipitation...''). To accommodate both forms, we provide two versions of the seeding statement. Step \Circled{0} in Listing~\ref{lst: gen_prompt} shows the version for action-based LOs; the version for content-based learning objectives is similar, with minor wording adjustments in bold: ``\texttt{The exercises below are designed to test whether a student \textbf{understands the following facts:} \{content-based LO\}}''. 

The first step involving actual text generation is to create a batch of question stems that all correspond to the given LO. We extend the seeding statement with a directive asking Phi-2 to create an MCQ that has the specified number of answer choices, as Step \Circled{1} shows. To create diverse and plausible question stems, we use nucleus sampling~\cite{holtzman2020}, with a temperature of $1.0$ and a \texttt{top\_p} value of $0.95$, to generate a predefined number of question stems in each batch. We repeat this process through multiple batches until we reach the target number of questions per LO. 

After the question stems are generated, they become part of the prompt for the next step, which is to generate the answer choices in the case of MCQ generation. Step \Circled{2} in Listing~\ref{lst: gen_prompt} shows the prompt used to generate answer choices. We use the choice labels ``\texttt{a)}'', ``\texttt{b)}'' and so on as anchors to guide Phi-2 to generate the answer choices in a structured format. Answer choices are generated one at a time, with the prompt extended to include all previously generated choices---for example, when generating the choice \texttt{c)}, the prompt will include the question stem and the choices \texttt{a)} and \texttt{b)}. This incremental prompt completion strategy helps maintain consistency among the answer choices. To maximize the quality of the generated answer choices, we use a beam search of width 2 to assist in the generation process, which allows us to explore multiple plausible continuations while ensuring that the generated choices are coherent and relevant to each question stem. All question stems in a batch get their answer choices in parallel.

The next step is to generate the correct answer for each question. For an MCQ where a single correct answer is expected, this means selecting one of the answer choices based on their probabilities. We extend the previous prompt with another prompt completion step, as shown in Step \Circled{3}, to ask Phi-2 to select the correct answer. We then use \emph{greedy decoding} (\texttt{top\_k} = 1) to identify the most probable choice label and copy the corresponding answer choice verbatim to complete the prompt. This ensures that the generated answer is always one of the provided answer choices. 

The final step in the generation stage is to generate an explanation for the correct answer. Since we do not expect the explanation to vary significantly (a choice can only be correct for a limited number of reasons), we use greedy decoding (without random sampling) to complete the prompt for explanation as shown in Step \Circled{4}, and limit the maximum length of the generated explanation to 200 tokens.

\begin{table*}[t]
\centering
\small
\caption{A list of exclusion criteria used in syntactic filtering and their rationales}
\label{tab: filtering}
\begin{tabular}{|l|p{6cm}|p{6cm}|}
\hline
\textbf{Category} & \textbf{Exclusion Criteria} & \textbf{Rationale} \\
\hline
Stem & Has fewer than 10 characters & The average English word length is about 5; 10 characters are no more than two informative words.  \\
\hline
\multirow[t]{5}{*}{Answer Choices} & Have empty or duplicate choices & These signal an incomplete or malformed question. \\
\cline{2-3} % \cline only draws a line over the specified columns
& Have fewer than 5 characters & This suggests a trivial choice. \\
\cline{2-3}
& Include flawed choices: ``all of the above'' or ``none of the above'' & These are clear item-writing flaws~\cite{Tarrant2006}. Students know these are usually the correct choice.\\
\cline{2-3}
& Include dichotomous choices: ``yes'', ``no'', ``true'', or ``false'' & These are indicative of a degenerate MCQ that is actually a true/false question. \\
\cline{2-3}
& Start with ``both'' or ``neither'' & These are indicative of an ineffective K-type MCQ~\cite{Tarrant2006} asking students to choose an answer combination. \\
\hline
Explanation & Has fewer than 10 characters & This suggests an inadequate explanation, which may arise from a systemic flaw in the question. \\
\hline
\end{tabular}
\end{table*}

\subsection{Stage 2: Selective Validation}\label{sec: validation}

After the four incremental generation steps, we have created an abundance of MCQs, each having a stem, multiple answer choices, an answer, and an explanation. However, we have not yet reviewed the quality of these questions. Some may contain a trivial stem, include incomplete answer choices, or \emph{actually} have no correct answer. In fact, the expansive generation stage operates in batch processing mode to maximize efficiency; any intermediate quality checks would disrupt the flow and slow down the overall process. Therefore, we devote a separate stage in our pipeline to validating the generated questions and removing low-quality ones.

A low-quality MCQ is likely to contain one or more item-writing flaws~\cite{Tarrant2006}, such as including ``all of the above'' as an answer choice or having no single best answer, which violate commonly accepted MCQ design guidelines~\cite{haladyna2004}. Since our pipeline generates questions based on a given LO, we also want to ensure that each question indeed assesses students on its intended LO. To select high-quality questions, we implement three validation steps of increasing sophistication: (1) \textbf{syntactic filtering} based on textual features, (2) \textbf{answer confidence evaluation} based on probabilistic reasoning, and (3) \textbf{learning-objective alignment check} based on re-classification.

\subsubsection{Syntactic Filtering}\label{sec: filtering}

We perform a simple rule-based filtering to remove questions with obvious syntactic flaws. This includes, for example, de-duplicating questions with identical stems and answer choices, eliminating trivial questions with exceedingly short stems (fewer than 10 characters) or answers (fewer than 5 characters), and removing questions with indicative answer choices such as ``all of the above'' or ``none of the above''. Table~\ref{tab: filtering} details all the exclusion criteria we used for syntactic filtering and their rationales.

\begin{figure*}[t]
\setcounter{lstlisting}{1}
\begin{center}
\begin{minipage}[t]{\textwidth}
\captionof{lstlisting}{The prompt template used to extract choice probabilities (\textbf{left}) and a concrete example (\textbf{right}).}
\label{lst: val_prompt}
\begin{minipage}[t]{0.48\textwidth}
    \begin{lstlisting}[frame=single,basicstyle=\ttfamily\small,showspaces=false,escapechar=\%]
%\textbf{Exercise 1:}%
%\textbf{Multiple Choice:}%
{
    stem
}
{
    original choices
}
%\textbf{e) None of the above}%
%\textbf{Answer:}%
[choice probabilities]
\end{lstlisting}
\end{minipage}\hfill
\begin{minipage}[t]{0.48\textwidth}
    \begin{lstlisting}[frame=single,basicstyle=\ttfamily\small,showspaces=false,escapechar=\%]
%\textbf{Exercise 1:}%
%\textbf{Multiple Choice:}%
Which of the following factors contributes 
most to the degradation of mangroves?
a) Climate change
b) Overfishing
c) Pollution
d) Urbanization
%\textbf{e) None of the above}%
%\textbf{Answer:}% [a: 0.28, b: 0.20, c: 0.13, 
        d: 0.24, e: 0.15]
    \end{lstlisting}
\end{minipage}
\end{minipage}
\end{center}
\end{figure*}

\begin{figure*}[t]
\setcounter{lstlisting}{2}
\begin{center}
\begin{minipage}[t]{\textwidth}
\captionof{lstlisting}{Example prompts asking Phi-2 to evaluate $\log \Pr(Q | L)$ (\textbf{left}) or $\log \Pr(Q)$ (\textbf{right}) for a fixed question $Q$ but a variable learning objective $L$ (enclosed by the curly braces \{\}). In both cases Phi-2 evaluates the log-probability of the same question text, but for $\log \Pr(Q | L)$ on the left, the question text is prefixed by another piece of text that introduces the learning objective. }
\label{lst: rel_prompt}
\begin{minipage}[t]{0.48\textwidth}
    \begin{lstlisting}[frame=single,basicstyle=\ttfamily\small,showspaces=false,escapechar=\%]
%\textbf{The exercise below is designed to test whether}%
%\textbf{a student can}% 
{describe different methods of irrigation}%\textbf{.}%

%\textbf{Multiple Choice:}%
Which irrigation method provides the most 
water directly to the roots?
a) Drip irrigation
b) Sprinkler irrigation
c) Furrow irrigation
d) Flood irrigation
Answer: a
\end{lstlisting}
\end{minipage}\hfill
\begin{minipage}[t]{0.48\textwidth}
    \begin{lstlisting}[frame=single,basicstyle=\ttfamily\small,showspaces=false,escapechar=\%]
      ## Intentionally left blank 
          
          to provide no context  ##

%\textbf{Multiple Choice:}%
Which irrigation method provides the most 
water directly to the roots?
a) Drip irrigation
b) Sprinkler irrigation
c) Furrow irrigation
d) Flood irrigation
Answer: a
    \end{lstlisting}
\end{minipage}
\end{minipage}
\end{center}
\end{figure*}

\subsubsection{Answer Confidence Evaluation}\label{sec: ans_conf}

The generation stage uses Phi-2 to create questions and also generate the answers. But, \textbf{being an SLM proficient in math and science, how confident is Phi-2 about its own answers?} If Phi-2 is uncertain about the answer to a question it generated, this inevitably raises doubts about the quality of the question itself---maybe the question is ambiguous or poorly constructed, or maybe the answer choices are not well designed.
We introduce a novel validation step that leverages Phi-2's \textbf{probabilistic reasoning ability} to evaluate its confidence in its generated answers and remove questions that even Phi-2 is uncertain about how to answer.

For each question that has passed syntactic filtering, we first shuffle the order of its answer choices to mitigate the position bias where Phi-2 may unevenly choose a particular position (e.g., the first choice) to generate the correct answer. Notably, this shuffling of choices is straightforward because we do not have answer choices such as ``both \texttt{a} and \texttt{b}'' or ``none of the above'', which would otherwise create interdependent answer choices that need careful handling.

However, to allow Phi-2 to indicate that none of the answer choices provided is correct, we \emph{add the ``none of the above'' option back} to every question after shuffling the original choices, just for this validation step. This way, we can identify potential low-quality questions for which Phi-2 does not reject the possibility that no correct answer is present. We could have separately added the ``all of the above'' option to detect an ``all-correct'' case, but this detection is already built into our approach based on probabilistic reasoning, which we will describe next.

We embed each question in the prompt template shown in Listing~\ref{lst: val_prompt}, which asks Phi-2 to answer the question. The most natural next token to complete this prompt is one of the choice labels (e.g., \texttt{c}), and since Phi-2 is open-source, we can extract the next-token probabilities for these labels (the original choices \texttt{a}-\texttt{d} plus the extra choice \texttt{e} for ``none of the above''). By applying the \texttt{softmax} function, we can obtain a properly normalized probability distribution over the choice labels, which reflects Phi-2's confidence in each answer choice being correct. 

Our goal is to select questions that Phi-2 determined to have an unambiguous answer with high confidence. Therefore, we prefer questions whose choice distribution has an outstanding peak, indicating that Phi-2 is disproportionately more confident about one particular choice over the others. If the choice distribution is flat and close to uniform, the question may be ambiguous or have ``all of the above'' as the correct answer, since Phi-2 is equally confident about all choices. We also want to ensure that the generated answer, rather than ``none of the above'' option, leads the other choices in probability \emph{by a significant margin}. Therefore, we only select questions whose most probable choice has a probability greater than a high threshold (e.g., 90\%) and is not the ``none of the above'' option.

\subsubsection{Learning-Objective Alignment Check}\label{sec: alignment}

The five-step incremental prompt shown in Listing~\ref{lst: gen_prompt} guides Phi-2 to generate a MCQ from a given LO. But \textbf{does the generated question most closely align with its original LO rather than some other LOs?} A high-quality question should be specifically relevant to the LO it is intended to assess, in addition to having a solid structure and an unambiguous answer. Aligning better with a different LO than the original may indicate that the question is flawed or at least not optimal for the intended purpose. To assess each question's alignment with its original LO, we repurpose Phi-2 as a classifier that ranks all the LOs for each question in descending order of relevance, and we select only those questions whose top-1 identified LO matches the original.

Inspired by recent work that uses question co-occurrence probabilities to identify related questions~\cite{wei2025}, we developed a novel measure of relevance between a question and an LO based on the change in question log-probability when the LO is included in the context. Specifically, we define the relevance of a learning objective $L$ to a question $Q$ as the difference in the log-probability of $Q$ when $L$ is present in the context versus when it is absent:
\begin{equation}
    \text{Relevance}(L, Q) = \log \Pr(Q | L) - \log \Pr(Q)
\end{equation}
where $\Pr(Q | L)$ is the probability of observing the question $Q$ when the learning objective $L$ is included in the context, and $\Pr(Q)$ is the probability of observing $Q$ in a neutral context without any specific learning objective.

If including $L$ in the context significantly increases the probability of $Q$, it suggests that $L$ is highly relevant to $Q$; or, if the probability remains unchanged or decreases, $L$ and $Q$ are unlikely to have a strong connection. We use the algorithm described in previous work~\cite{wei2025} to compute these log-probabilities using Phi-2 and construct two prompt templates. The first template, an example of which is shown on the left of Listing~\ref{lst: rel_prompt}, is used to compute $\Pr(Q | L)$ by prefixing the question $Q$ with a learning objective $L$. In the other template used to compute $\Pr(Q)$, however, we only include the question $Q$ but provide an empty context. We run the algorithm for every pair of LO and question, producing a rectangular relevance matrix where each row corresponds to an LO and each column a question. We then determine for each question which LO has the highest relevance score and only retain a question if the most relevant LO matches the original LO used to generate the question.

\begin{table*}[t]
\centering
\small
\caption{The list of eight learning objectives randomly selected to use in the evaluation studies, along with the unit and topic in which they are included. There are nine units in total in the AP Environmental Science Course and Exam Description. These selected learning objectives cover eight distinct topics and almost half of the available units.}
\label{tab: lo}
\begin{tabular}{|p{1.5cm}|p{4cm}|p{8.5cm}|}
\hline
\textbf{Unit} & \textbf{Topic} & \textbf{Learning Objective} \\
\hline
% \parbox{1.5cm}{\vspace{.3\baselineskip} \centering Unit 1: The Living World: Ecosystems \vspace{.2\baselineskip}} 
\multirow{2}{*}{\parbox{1.5cm} {\centering (Unit 1) Ecosystems}} & (Topic 1.4) The Carbon Cycle & (ERT-1.D) Explain the steps and reservoir interactions in the carbon cycle. \\
& & \\
\hline
\multirow{2}{*}{\parbox{1.5cm}{\centering (Unit 3) Populations}} & (Topic 3.4) Carrying Capacity & (ERT-3.E) Describe the impact of carrying capacity on ecosystems. \\
 \cline{2-3}
& (Topic 3.9)
Demographic
Transition & (EIN-1.D) Define the demographic transition. \\
 % \cline{1-3}
\hline
\multirow{3}{*}{\parbox{1.5cm}{\centering (Unit 5) \\ Land and Water Use}} & (Topic 5.8) Impacts of Overfishing & (EIN-2.J) Describe causes of and problems related to overfishing. \\
\cline{2-3}
& (Topic 5.15) Sustainable Agriculture & (STB-1.E) Describe sustainable agricultural and food production practices. \\
\cline{2-3}
& (Topic 5.17) Sustainable Forestry & (STB-1.G) Describe methods for mitigating human impact on forests.\\
% \cline{1-3}
\hline
\multirow{3}{*}{\parbox{1.5cm}{\centering (Unit 8) Aquatic and Terrestrial Pollution}} & (Topic 8.4)
Human Impacts on Wetlands and Mangroves & (STB-3.E) Describe the impacts of human activity on wetlands and mangroves.\\
\cline{2-3}
& (Topic 8.5) Eutrophication & (STB-3.F) Explain the environmental effects of excessive use of fertilizers and detergents on aquatic ecosystems. \\
\hline
\end{tabular}
\end{table*}

\section{Evaluation Studies}\label{sec: evaluation}

Evaluating automatically generated questions is a challenging task. At a minimum, a well-written question should have an unambiguous answer and be relevant to the LO that it is intended to test~\cite{haladyna2004}. We designed two evaluation studies to assess these two aspects of the questions generated by our pipeline: \textbf{a human evaluation study} involving middle-school science teachers as expert judges and \textbf{an automatic evaluation study} using an advanced LLM as a surrogate judge. Both studies used the same test items and followed the same experimental design.

We collected 41 action-based LOs from the Advanced Placement (AP) Environmental Science Course and Exam Description\footnote{\url{https://apcentral.collegeboard.org/media/pdf/ap-environmental-science-course-and-exam-description.pdf}} and used our question generation pipeline to generate 200 MCQs for each LO, creating an expansive candidate pool of 8,200 questions. After syntactic filtering (Section~\ref{sec: filtering}), we retained 6,043 questions (73.7\% of the original pool). Next, with a probability threshold of $0.9$, we performed answer confidence evaluation (Section~\ref{sec: ans_conf}) and refined the pool to only include 3,426 questions. Lastly, the learning-objective alignment check (Section~\ref{sec: alignment}) produced a final pool of 3,205 questions covering all 41 LOs. The average number of MCQs per learning objective was 78.17 (SD = 29.05). We \emph{randomly} selected eight LOs, each of which had at least eight questions, creating an evaluation set of 64 MCQs. The selected learning objectives are listed in Table~\ref{tab: lo}.

\subsection{Study 1: Humans as Expert Judges}
We recruited seven middle-school science teachers to participate in our human evaluation study. The study was conducted online through a Qualtrics survey, in which every participant received the same set of 64 MCQs. For each MCQ being evaluated, the participants were asked to perform two tasks displayed on a single page, which corresponded to the first two research questions (RQ-1 and RQ-2) that we aimed to investigate. 

First, they were asked to answer the MCQ themselves by selecting one of the provided answer choices as if they were students taking a test. Although after the syntactic filtering step, none of the MCQs would contain the trivial ``none of the above'' option, we \emph{added this option back} for this task as we did with answer confidence evaluation, so that the participants could indicate that an MCQ is unanswerable. After that, they moved on to the second task. Resuming their role as expert science teachers, they were asked to determine, by selecting ``Yes'' or ``No'', whether the MCQ they had just answered tests students on a given LO. As part of the experimental manipulation, the LO shown to the teachers was either the one used to generate the MCQ (the control condition) or a different LO randomly selected from the other seven learning objectives (the treatment condition). The right half of Figure~\ref{fig: gemini} shows the two tasks that participants saw during the experiment. 

We maintained a balanced, within-subject design: all participants evaluated the same MCQs in both conditions, and each condition had 32 questions covering all eight learning objectives (four questions each). To mitigate order effects, the order of the questions and conditions was randomized for each participant. Participants were not informed about the experimental manipulation during the consent process or the study itself to avoid biasing their judgments. Participants were recruited from a larger sample of science teachers participating in one of our projects. They were compensated with a \$25 gift card for 30 minutes of their time. All procedures were approved by the appropriate IRB.

\setcounter{figure}{0}
\begin{figure*}[t]
    \centering \includegraphics[width=\textwidth]{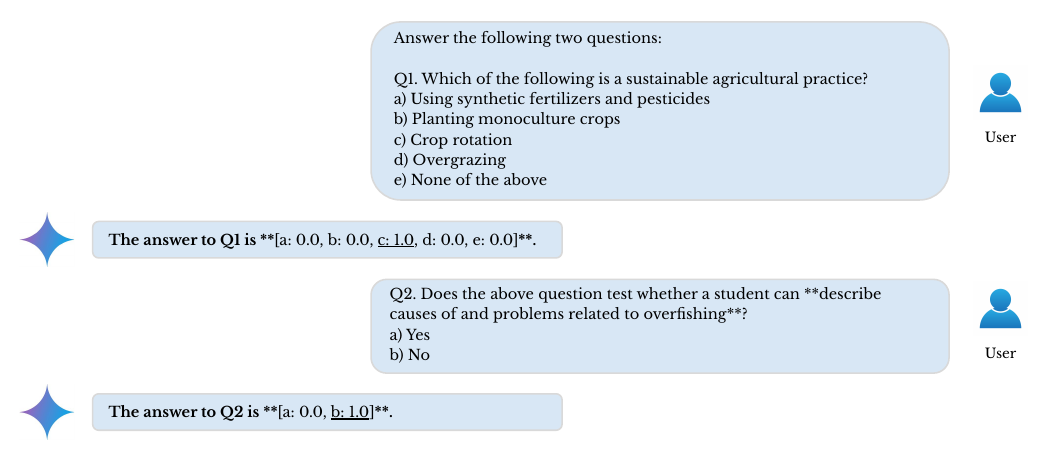}
    \caption{A visual illustration of a two-turn conversation we had with Gemini to simulate human survey experience. We prefixed Gemini's response with a scaffold (in bold) to force it to assign high probabilities to the choice labels. The label with the highest probability (underlined) is used to complete Gemini's response. Human participants saw the exact same text that ``User'' is verbalizing.}
    \label{fig: gemini}
    \Description{This image depicts a two-turn conversation with Gemini. In the first turn, the user provides instructions and a multiple-choice question, and Gemini responds with the answer choice it deems correct. In the second turn, the user asks whether the question aligns with a given learning objective, and Gemini responds with either "Yes" or "No".}
\end{figure*}

\subsection{Study 2: LLM as a Surrogate Judge}

To complement our human evaluation study, we also conducted an automatic evaluation study using an advanced LLM as a surrogate judge to assess the quality of the generated MCQs. This approach aligns with the recent research on ``LLM-as-a-Judge''~\cite{zheng2023,gu2024} that explores the use of LLMs to evaluate the outputs of other LLMs or automated systems. We used Gemini-2.5-Pro~\cite{google2025} (hereinafter referred to as ``Gemini''), a state-of-the-art LLM with strong reasoning capabilities developed by Google DeepMind, as our surrogate judge to answer our RQ-3 and RQ-4.  

Gemini was prompted to evaluate the same 64 MCQs under the same experimental design as the human evaluation study. To simulate the human survey experience, we presented the evaluation instructions seen by human participants verbatim to Gemini in a two-turn conversation with scaffolds as shown in Figure~\ref{fig: gemini}. In the first turn, we initiated the conversation by setting a question-answering context and asking Gemini to answer a generated MCQ. Instead of expecting a free response, we prefixed the model's response with a leading statement, ``The answer to Q1 is **'', which forced the model to assign more probability mass to the choice labels than other tokens. We then extracted the next-token probabilities for each label and selected the most probable one to complete the model's response, similar to how we evaluated Phi-2's answer confidence. In the second turn, we asked Gemini the same survey question answered by human participants, and again prefixed its response with a leading statement to extract the next-token probabilities for ``Yes'' and ``No''. Each MCQ was evaluated independently in a separate two-turn conversation to avoid carry-over effects. The automatic evaluation took seconds to complete and cost less than \$1 in API usage fees.

\section{Results and Discussion}

Our four research questions focus on the agreement between and among different judges: human experts, Phi-2, and Gemini. Given that these judges evaluated nominal variables (answer choices and yes/no), we used standard inter-rater reliability metrics~\cite{mchugh2012}, including Cohen's kappa (for two raters)~\cite{cohen1960} and Fleiss' kappa (for multiple raters)~\cite{fleiss1971}, to quantify the level of inter-rater agreement.

\setcounter{figure}{1}
\begin{figure*}[t]
     \centering
     \begin{subfigure}[t]{0.48\textwidth}
         \centering
\includegraphics[width=\textwidth]{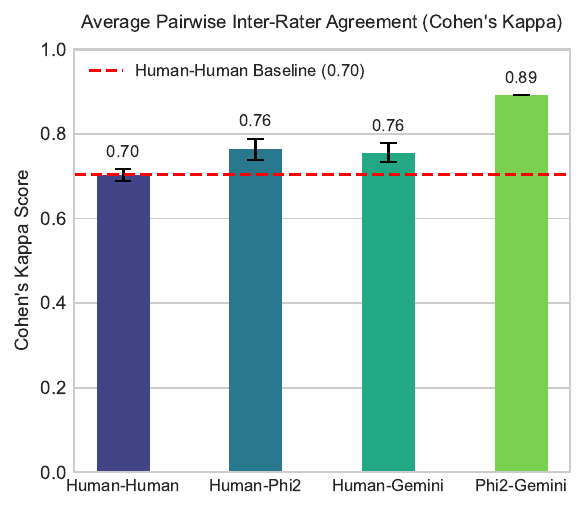}
         \caption{Average pairwise Cohen's kappa}
         \label{fig: q1_cohen}
         \Description{A bar chart showing the average pairwise Cohen's kappa values for Human vs. Human, Human vs. Phi-2, Human vs. Gemini, and Phi-2 vs. Gemini. The values are approximately 0.7, 0.76, 0.76, and 0.89, respectively.}
     \end{subfigure}
     \hfill
     \begin{subfigure}[t]{0.46\textwidth}
         \centering
         \includegraphics[width=\textwidth]{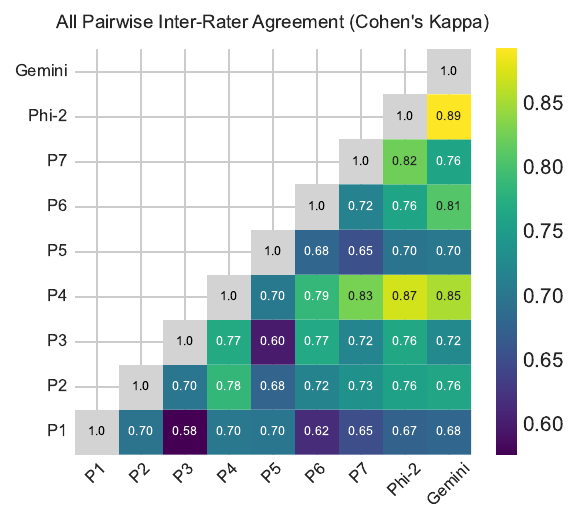}
         \caption{All pairwise Cohen's kappa (``P'' for human judges)}
         \label{fig: q1_heatmap}
         \Description{A heatmap showing the pairwise Cohen's kappa values between all distinct judges, including seven human judges (P1 to P7), Phi-2, and Gemini. The values range from approximately 0.6 to 0.9, with most values being above 0.7, indicating substantial agreement among the judges.}
     \end{subfigure}
    \caption{Inter-rater agreement on correct answers (RQ-1 and RQ-3)}
    \label{fig:three graphs}
    \Description{This figure contains two subfigures. The first subfigure is a bar chart showing the average pairwise Cohen's kappa values between different judges and the second subfigure is a heatmap showing the pairwise Cohen's kappa values between all pairs of distinct judges.}
\end{figure*}

\subsection{Agreement on Correct Answers (RQ-1 and RQ-3)}\label{sec: results_rq1}

We first examine the agreement on the correct answers to the 64 generated MCQs. Each MCQ had five answer choices (four original choices plus ``none of the above''), and each judge selected one choice as the answer. The seven human judges achieved a substantial agreement~\cite{landis1977} with each other (Fleiss' kappa = 0.70), indicating that \textbf{the generated MCQs are generally answerable and unambiguous}. Including either Phi-2 or Gemini as an eighth judge slightly increased the agreement (Fleiss' kappa = 0.72 in both cases), suggesting that \textbf{both models' judgments were generally consistent with those of the human experts}. 

Figure~\ref{fig: q1_cohen} shows the average pairwise Cohen's kappa values between two distinct judges (either human or machine). On average, the agreement between each pair of human judges was substantial (mean Cohen's kappa = 0.7, SE = 0.01), consistent with what Fleiss' kappa indicated. The pairwise agreement between Phi-2 and human judges was higher than that between human judges (mean Cohen's kappa = 0.76, SE = 0.03), suggesting that \textbf{Phi-2's answers were even more aligned with human experts than the experts were with each other}. Gemini achieved the same level of agreement with human judges (mean Cohen's kappa = 0.76, SE = 0.02) as Phi-2 did, indicating that Gemini's answers were also consistent with human judgments. The agreement between Phi-2 and Gemini was almost perfect (Cohen's kappa = 0.89), showing that \textbf{the two models largely concurred on the correct answers to the generated MCQs}. Figure~\ref{fig: q1_heatmap} presents a heatmap of the Cohen's kappa values between all pairs of distinct judges, visually confirming the patterns observed in Figure~\ref{fig: q1_cohen}. We did not observe any significant outliers among the judges, as all pairs of judges achieved a similar level of agreement. 

If we consider the majority vote among the seven human judges as the ground truth given by a single authoritative judge, \textbf{both Phi-2 and Gemini achieved almost perfect agreement with this majority judge}: Phi-2 had a Cohen's kappa of 0.87, while Gemini had a slightly smaller Cohen's kappa of 0.85. In terms of each model's raw accuracy in identifying the correct answers, Phi-2 had an accuracy of 90.6\% (58 out of 64), while Gemini had an accuracy of 89.1\% (57 out of 64). Both models made the same incorrect predictions on the same four MCQs.

There were 11 MCQs for which at least one human judge selected ``none of the above'', but majority voting among the human judges indicated that 10 of these MCQs \emph{had} a correct answer among the original choices. Only one MCQ (1.6\% of the 64 MCQs) was deemed unanswerable by the majority of human judges. For the six MCQs where only one human judge chose ``none of the above'', both Phi-2 and Gemini agreed with the rest of human judges on the correct answer, suggesting that one human judge might have slipped or misread the question. 

Overall, these results suggest that \textbf{the MCQs generated by our pipeline were clear and had unambiguous answers}, as evidenced by the substantial agreement among human judges on the correct answers, as well as the strong alignment between the two language models and the human judges. The high agreement between human judges and Gemini also indicated that Gemini can effectively replicate human judgments in answering MCQs generated from our pipeline, highlighting its potential as a reliable surrogate judge for the expansive generation stage of our pipeline.

\subsection{Agreement on Learning-Objective Alignment (RQ-2 and RQ-4)}

\setcounter{figure}{2}
\begin{figure*}[t]
     \centering
     \begin{subfigure}[t]{0.48\textwidth}
         \centering
\includegraphics[width=\textwidth]{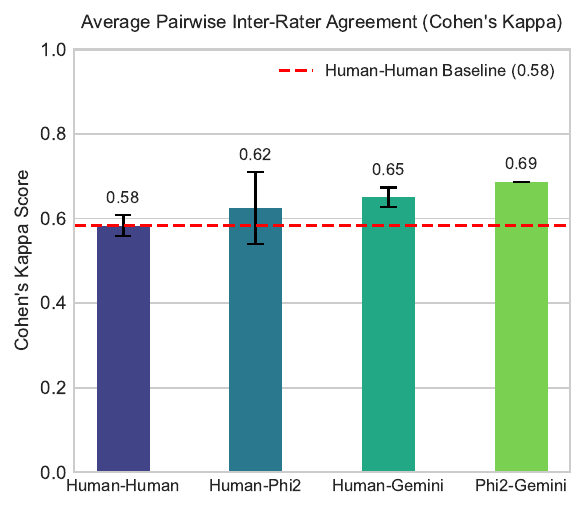}
         \caption{Average pairwise Cohen's kappa}
         \label{fig: q2_cohen}
         \Description{A bar chart showing the average pairwise Cohen's kappa values for Human vs. Human, Human vs. Phi-2, Human vs. Gemini, and Phi-2 vs. Gemini. The values are approximately 0.58, 0.62, 0.65, and 0.69, respectively.}
     \end{subfigure}
     \hfill
     \begin{subfigure}[t]{0.45\textwidth}
         \centering
         \includegraphics[width=\textwidth]{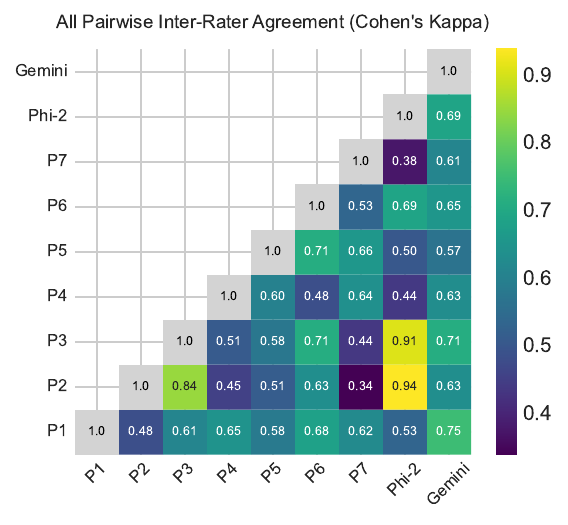}
         \caption{All pairwise Cohen's kappa (``P'' for human judges)}
         \label{fig: q2_heatmap}
         \Description{A heatmap showing the pairwise Cohen's kappa values between all distinct judges, including seven human judges (P1 to P7), Phi-2, and Gemini. The values range from approximately 0.4 to 0.9, with most values being above 0.5, indicating moderate agreement among the judges.}
     \end{subfigure}
    \caption{Inter-rater agreement on learning-objective alignment (RQ-2 and RQ-4)}
    \Description{This figure contains two subfigures. The first subfigure is a bar chart showing the average pairwise Cohen's kappa values between different judges and the second subfigure is a heatmap showing the pairwise Cohen's kappa values between all pairs of distinct judges.}
\end{figure*}

\setcounter{figure}{3}
\begin{figure*}[t]
     \centering
     \begin{subfigure}[t]{0.3\textwidth}
         \centering
\includegraphics[width=\textwidth]{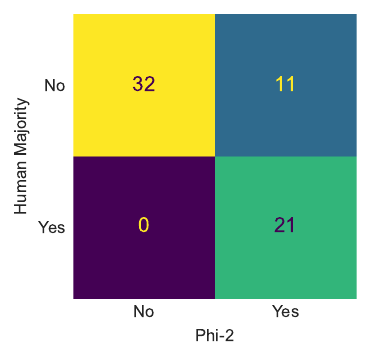}
         \caption{Human Majority vs. Phi-2}
         \label{fig: q2_cm_human_phi2}
     \end{subfigure}
     \hfill
     \begin{subfigure}[t]{0.3\textwidth}
         \centering
         \includegraphics[width=\textwidth]{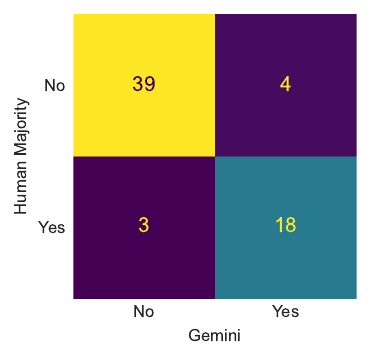}
         \caption{Human Majority vs. Gemini}
         \label{fig: q2_cm_human_gemini}
     \end{subfigure}\hfill
     \begin{subfigure}[t]{0.3\textwidth}
         \centering
\includegraphics[width=\textwidth]{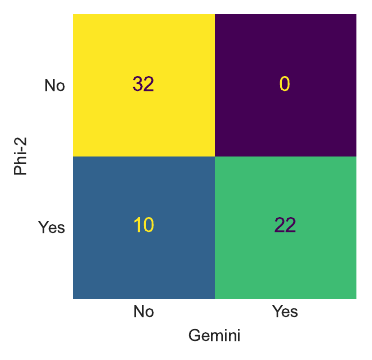}
         \caption{Phi-2 vs. Gemini}
         \label{fig: q2_cm_phi2_gemini}
     \end{subfigure}
        \caption{Three confusion matrices showing the number of agreements and disagreements between different judges}
        \Description{This figure contains three confusion matrices. The first matrix compares the human majority with Phi-2, the second matrix compares the human majority with Gemini, and the third matrix compares Phi-2 with Gemini. Each matrix shows the counts of agreements and disagreements between the two judges being compared.} 
\end{figure*}

Next, we examine the agreement on whether a generated MCQ tests a given learning objective. Each judge (human or machine) made a binary judgment (Yes/No) for each MCQ regarding a specified learning objective. This seemed to be an inherently more challenging task than answering the MCQs, as it requires the judges to subjectively assess the relevance of each MCQ to an LO rather than objectively identifying the correct answer. The seven human judges achieved moderate agreement with each other (Fleiss' kappa = 0.58), indicating that \textbf{while there was some consensus among the human judges, there were also notable differences in their judgments regarding the alignment of MCQs with learning objectives}. Similar to what was observed in Section~\ref{sec: results_rq1}, including either Phi-2 or Gemini as an eighth judge slightly increased the agreement (Fleiss' kappa = 0.59), suggesting that \textbf{both models' judgments were generally consistent with those of the human judges}, although the overall agreement among judges was lower.

Figure~\ref{fig: q2_cohen} confirmed this observation of lower agreement among judges. The average pairwise Cohen's kappa between two distinct human judges indicated a moderate agreement (mean Cohen's kappa = 0.58, SE = 0.02). The pairwise agreement between Phi-2 and human judges was slightly above the conventional threshold (0.6) of being substantial (mean Cohen's kappa = 0.62, SE = 0.08), suggesting that \textbf{Phi-2's judgments were somewhat aligned with human experts but with more variability than in the first task}. Gemini achieved slightly better alignment with human judges (mean Cohen's kappa = 0.65, SE = 0.02). Similar to the first task, Phi-2 and Gemini still attained the best agreement (Cohen's kappa = 0.69) among all pairs of judges, \textbf{indicating that the two models were more consistent with each other than with human judges}. 

Figure~\ref{fig: q2_heatmap} shows a heatmap of the Cohen's kappa values between all pairs of distinct judges. We noticed that two human judges, P2 and P3, had exceptionally high agreement with Phi-2 (Cohen's kappa = 0.94 and 0.91, respectively), but they achieved only moderate agreement with other human judges, suggesting that these two judges might have had a similar interpretation of the learning objectives as Phi-2 did. This also highlighted the inherent subjectivity in judging learning-objective alignment, as different experts may have varying interpretations of what constitutes alignment. 

If we again consider the human majority as the ground truth, we can visualize the number of agreements and disagreements between two judges in a $2 \times 2$ confusion matrix, since determining LO alignment is a binary classification task. Figure~\ref{fig: q2_cm_human_phi2} shows the confusion matrix between the human majority and Phi-2. We observed that the human majority identified all the mismatched LOs in the treatment condition (Phi-2 = ``No''), but showed 11 disagreements with Phi-2 in the control condition (Phi-2 = ``Yes''). Gemini showed a closer alignment with the human majority, as evidenced by a more balanced confusion matrix against the human majority (Figure~\ref{fig: q2_cm_human_gemini}); it also had a confusion matrix against Phi-2 (Figure~\ref{fig: q2_cm_phi2_gemini}) that was similar to that between the human majority and Phi-2. 

\setcounter{figure}{4}
\begin{figure*}[t]
    \centering
\includegraphics[width=0.85\textwidth]{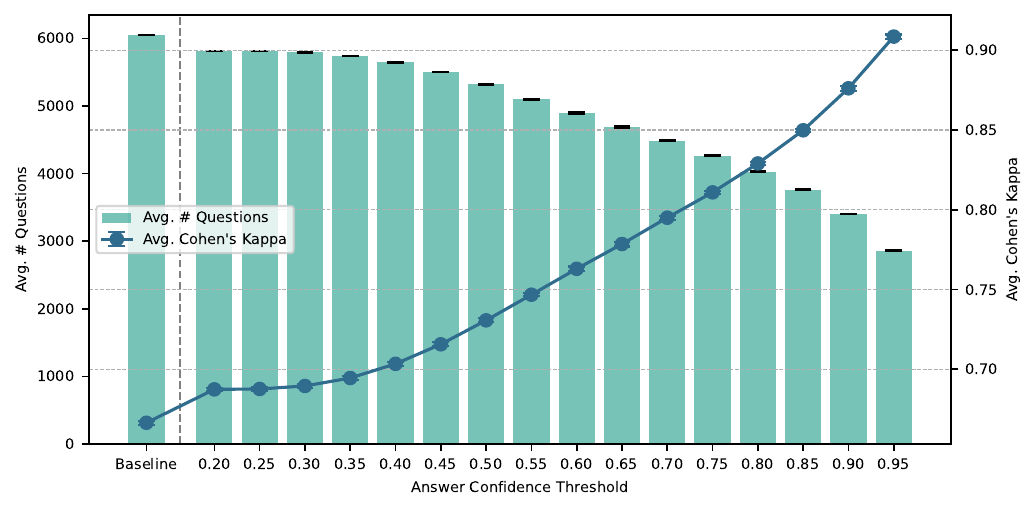}
\caption{A higher answer confidence threshold reduces the number of retained MCQs but improves their quality}
\label{fig: ans_conf}
\Description{This figure is a line chart showing the relationship between the answer confidence threshold and two metrics: the average number of retained MCQs and the average Cohen's kappa between Phi-2 and Gemini. As the answer confidence threshold increases from baseline to 95\%, the average number of retained MCQs decreases from 6043 to 2856, while the average Cohen's kappa increases from 0.67 to 0.91.}
\end{figure*}

Overall, these results suggested that while the MCQs generated by our pipeline were generally relevant to their intended LOs, there was more variability in judgments about LO alignment compared to answering the MCQs. The moderate agreement among human experts indicated that assessing LO alignment is a more subjective task, with different experts potentially having diverse interpretations of what constitutes alignment. The slightly better alignment of Gemini with human judges compared to Phi-2 suggests that Gemini may have a better understanding of the nuances in different LOs, making it a reasonable surrogate judge for evaluating LO alignment.

\subsection{Ablation Study}

When describing the design of our evaluation studies in Section~\ref{sec: evaluation}, we brought up a seemingly arbitrary choice of using a probability threshold of 90\% in the answer confidence evaluation (Section~\ref{sec: ans_conf}). This choice was based on our intuition that a high-confidence threshold would help remove low-quality questions, but we did not have empirical evidence to support this intuition. More fundamentally, are the two elaborate validation steps after syntactic filtering, namely, answer confidence evaluation and LO alignment check, really necessary? Could we have achieved similar results with only syntactic filtering? These are important questions to investigate as they can reinforce or challenge our design choices and inform future improvements. To better understand how each step of the validation stage had contributed, we conducted an ablation study. We created three additional evaluation sets by progressively applying the three validation steps to the original candidate pool of 8,200 generated MCQs:
\begin{itemize}[leftmargin=*]
    \item \textbf{Version A (Step 1)}: This version includes the 6,043 MCQs that passed only the \emph{syntactic filtering} step.
    \item \textbf{Version B (Step 1 + Step 2)}: This version includes a subset of Version A that also passed the \emph{answer confidence evaluation} step at varying probability thresholds, from 20\% to 95\% in increments of 5 percentage points.
    \item \textbf{Version C (Step 1 + Step 2 + Step 3)}: This version includes a subset of Version B that also passed the \emph{learning-objective alignment check} step, again at varying probability thresholds from 20\% to 95\%.
\end{itemize}

We considered all 41 learning objectives that we had available. Since all three versions involved some degree of randomness (e.g., in shuffling answer choices or GPU computation), we repeated the ablation study ten times with different random seeds to ensure the robustness of our findings. This would have created enormous workload for human judges, but since Gemini had shown a strong agreement with human judges in previous evaluation studies, we used Gemini as a surrogate judge to evaluate each version. 

Figure~\ref{fig: ans_conf} shows how the answer confidence threshold affected the number of retained MCQs and their quality as judged by Gemini. The horizontal axis represents the various probability thresholds used to create Version B, with ``Baseline'' indicating Version A (no answer confidence evaluation). The left vertical axis shows the average number of retained MCQs per ten runs, while the right vertical axis shows the average Cohen's kappa per ten runs measuring the agreement between Phi-2 and Gemini on the correct answers to the retained MCQs. As expected, a higher answer-confidence threshold implying a more stringent filtering process led to fewer retained MCQs. The average number of retained MCQs steadily decreased from 6,043 at the baseline (Version A) to about 2,856 at the 95\% threshold. This halving of retained MCQs, however, came with a significant improvement in quality. The average Cohen's kappa between Phi-2 and Gemini improved from 0.67 at the baseline to 0.91 at the 95\% threshold, indicating a substantial increase in agreement on the correct answers. This suggests that the answer confidence evaluation step was effective.

We also applied Gemini to evaluate the LO alignment between Version B and C. We observed a significant ($t(18) = 7.76, p < .001$) improvement in the percentage agreement between Phi-2 and Gemini on LO alignment, from 59\% at Version B to 61\% at Version C (Cohen's $d = 3.47$), at a 0.9 threshold. This indicates that the LO alignment check enhanced the relevance of the MCQs \emph{across all 41 LOs}, not just the 8 LOs used in the previous evaluation studies.

\section{Conclusion}
We have presented a novel pipeline for generating high-quality questions from learning objectives using an open-source SLM called Phi-2. Our pipeline adopts a novel ``generate-then-validate'' strategy, where it first generates an abundance of candidate questions through incremental prompt completion and then applies a series of validation steps to remove low-quality questions. We introduced two innovative validation techniques: answer confidence evaluation based on Phi-2's probabilistic reasoning ability and learning-objective alignment check based on re-classification.

Results from our comprehensive evaluation studies involving both human experts and Gemini show that our pipeline is effective in producing MCQs that have unambiguous answers and are generally relevant to their intended learning objectives. Our ablation study further demonstrates the necessity of our two novel validation techniques, as they significantly enhance the quality of the generated questions. 

One may share the concern that it seems wasteful to generate so many questions initially, only to discard a large portion of them during validation, as seen in Figure~\ref{fig: ans_conf}. However, we argue that \textbf{this is exactly how an SLM should be used: rapid generation at scale, followed by careful validation afterwards}. It is unrealistic to expect an SLM to succeed in one shot, but the law of large numbers will guarantee that good outcomes will emerge from an abundance of trials. Future work may explore a synergy between an SLM for generation and an LLM for validation, where the SLM performs fast generation and the LLM provides accurate feedback, both working together to deliver superior results.

\begin{acks}
This research was supported by the National Science Foundation under Grant No. 2301130 and a Google Academic Research Award to Paulo F. Carvalho.
\end{acks}

\bibliographystyle{ACM-Reference-Format}
\bibliography{main}

\end{document}